%% file: main.tex
\pdfoutput=1

\documentclass[11pt]{article}
\usepackage[final]{acl}

\usepackage{times}
\usepackage{latexsym}

\usepackage[T1]{fontenc}

\usepackage[utf8]{inputenc}

\usepackage{microtype}

\usepackage{inconsolata}

\usepackage{booktabs}
\usepackage{multirow}

\usepackage{graphicx}
\usepackage{amsthm}
\usepackage{amsmath}
\usepackage{bbm}

\usepackage{collectbox}
\usepackage{lipsum}  
\usepackage[frozencache, cachedir=minted-cache]{minted}

\usepackage{xcolor}

\usepackage[table]{xcolor}
\definecolor{LightGray}{gray}{0.9}
\definecolor{lightgreen}{RGB}{204,255,204}

\usepackage{algorithm}
\usepackage[noend]{algpseudocode}

\algrenewcommand\alglinenumber[1]{#1}

\usepackage{soul}

\title{Enhancing Foundation Models in Transaction Understanding with LLM-based Sentence Embeddings}

\author{Xiran Fan \quad Zhimeng Jiang \quad Chin-Chia Michael Yeh \\ \textbf{Yuzhong Chen \quad Yingtong Dou \quad Menghai Pan \quad Yan Zheng} \\
Visa Research \\ Foster City, CA, USA \\
 \{xirafan, zhimjian, miyeh, yuzchen, yidou, menpan, yazheng\}@visa.com}

\begin{document}
\maketitle
\begin{abstract}
\input{section/abstract}
\end{abstract}

\input{section/intro}

\input{section/relatedwork}

\input{section/methodology}

\input{section/experiments}

\input{section/conclusion}

\input{section/limitations}




\end{document}

%% file: section/abstract.tex
The ubiquity of payment networks generates vast transactional data encoding rich consumer and merchant behavioral patterns. Recent foundation models for transaction analysis process tabular data sequentially but rely on index-based representations for categorical merchant fields, causing substantial semantic information loss by converting rich textual data into discrete tokens. While Large Language Models (LLMs) can address this limitation through superior semantic understanding, their computational overhead challenges real-time financial deployment. We introduce a hybrid framework that uses LLM-generated embeddings as semantic initializations for lightweight transaction models, balancing interpretability with operational efficiency. Our approach employs multi-source data fusion to enrich merchant categorical fields and a one-word constraint principle for consistent embedding generation across LLM architectures. We systematically address data quality through noise filtering and context-aware enrichment. Experiments on large-scale transaction datasets demonstrate significant performance improvements across multiple transaction understanding tasks.

%% file: section/intro.tex
\section{Introduction}
\label{sec:intro}

Foundation models have achieved remarkable success across diverse domains, from natural language processing \cite{brown2020language, devlin2018bert} and computer vision \cite{dosovitskiy2020image,awais2025foundation} to multimodal learning \cite{ramesh2021zero,alayrac2022flamingo} and recommendation systems \cite{huang2024foundation}. Yet, their development for tabular data—prevalent in real-world applications—remains underexplored. Tabular data presents unique challenges, including permutation invariance, heterogeneous features, and domain-specific semantics that differ from text and images.

Recent work by \cite{zhang2023fata,yeh2025treasure} introduced a foundation model for tabular transactional data in the payment industry, showing promising results by leveraging sequential patterns across transactions. However, the model represents categorical merchant information by mapping names to discrete indices and learning embeddings, which leads to significant information loss. For example, simply mapping “Costco” to an index fails to capture its wholesale retail nature, membership-based model, or consumer behavior implications.

Large Language Models (LLMs), with their rich world knowledge and natural language understanding, offer an opportunity to bridge this semantic gap. Unlike index-based representations, LLMs can directly process textual information within transactions (e.g., merchant names and locations) to generate semantically meaningful embeddings. However, practical challenges arise in production environments, including computational and latency constraints, and the risk of introducing noise from real-world financial data.

We present a novel framework that bridges this gap by leveraging LLM-generated sentence embeddings to enhance foundation model performance while maintaining production viability. Our approach addresses the semantic impoverishment of categorical fields by using LLMs to generate contextualized embeddings that serve as initializations for foundation models. We focus on fields with enrichable contextual information as our primary use cases: merchant category codes (MCC), merchant names, and location information in transactional data. Our method employs multi-source data fusion to enrich these categorical fields and represents the enhanced information through carefully designed prompts optimized for consistent embedding generation.

A key innovation is our \textit{one-word limitation principle} \cite{jiang2024scaling} in prompt design, ensuring consistent and focused representations. For decoder-only models, we employ prompts such as ``This sentence: `[text]' means in one word:'' to generate focused outputs. We extract sentence embeddings from the final layer's hidden states, specifically selecting the last non-padding token's representation.

Our framework systematically addresses data quality through rule-based filtering and strategic null token replacement, while incorporating retrieval mechanisms to augment basic field information with relevant context. The resulting LLM-based embeddings initialize field representations in production models, enabling them to inherit semantic understanding while maintaining deployment efficiency through task-specific fine-tuning.

Our key contributions are: \textbf{(1)} A practical framework for integrating LLM-based sentence embeddings into foundation models that mitigates information loss in traditional index-based approaches; \textbf{(2)} A comprehensive preprocessing and prompt generation pipeline featuring the one-word limitation principle for consistent embedding generation; \textbf{(3)} A multi-source data fusion strategy for enriching categorical field information; \textbf{(4)} Empirical validation on real-world transaction data showing significant improvements in various  transaction metrics\footnote{Transaction volume, decline rate, and fraud rate are commonly used metrics for analyzing transactions in payment network companies.} across multiple LLM architectures.

%% file: section/relatedwork.tex
\section{Related Work}
\label{sec:related}

Our work is situated at the intersection of foundation models for tabular data, sequential transaction modeling, and semantic feature representation. 

\textbf{Foundation Models for Tabular Data} Foundation models, first popularized in NLP~\cite{brown2020language}, have recently been extended to tabular data~\cite{yang2023unitabe,van2024tabular,zhang2023towards}. Approaches like TabPFN~\cite{hollmann2022tabpfn} and tabular transformers mainly focus on static tables for entry classification, regression, and imputation, but are limited when applied to sequential transaction data where temporal dependencies matter.

\textbf{Sequential Transaction Modeling} Modeling transaction sequences requires capturing temporal patterns beyond isolated records. Recent work~\cite{skalski2023towards} addresses these temporal dynamics, supporting tasks such as transaction validation and next-transaction prediction. A notable contribution is FATA-Trans~\cite{zhang2023fata}, which addresses key limitations of earlier transformer-based models for sequential tabular data. Unlike approaches that replicate static fields across time steps and ignore temporal gaps, FATA-Trans employs separate transformers for static and dynamic fields, uses field-type embeddings to distinguish them, and applies time-aware position embeddings to capture both event order and time intervals. This design reduces redundancy, models temporal patterns effectively, and achieves state-of-the-art results with improved interpretability.

\textbf{Semantic vs. ID-based Feature Representation} Traditional tabular models use ID-based embeddings for categorical features, capturing collaborative filtering (CF) signals~\cite{schafer2007collaborative} but lacking semantic understanding and struggling with cold-start issues. Recent advances leverage LLMs to create semantic embeddings, and hybrid models (e.g., LLM2Rec~\cite{he2025large}) align CF and semantic signals for richer representations.

\textbf{LLMs for Structured Data} LLMs have been adapted to structured data tasks via serialization and fine-tuning, as in TableGPT2~\cite{su2024tablegpt2}. However, these methods mainly target general tabular tasks and require full LLM inference, which is impractical for large-scale, real-time transaction systems.

Building on principles introduced by FATA-Trans, our prior work~\cite{yeh2025treasure} scaled sequential tabular modeling to billions of real-world transaction records, creating a multi-purpose transformer-based foundation model for transaction understanding. This model processes transaction sequences and produces predictions for multiple downstream objectives—including transaction metric estimation and next-transaction prediction—through a shared multi-task output module with joint loss optimization. In this paper, we focus on optimizing the transaction encoding stage. In the original model, transaction vectors were formed by concatenating randomly initialized embeddings for each attribute (timestamp, amount, merchant name, MCC, location, abnormal labels). Here, we replace random initialization for selected categorical attributes (merchant name, MCC, location) with LLM-based semantic embeddings to capture richer contextual meaning, while preserving the remainder of the architecture and training procedure. Distinct from prior work, we address the challenge of sequential transaction modeling with a hybrid approach that integrates LLM-based semantic embeddings for categorical features while maintaining efficient production deployment. Unlike end-to-end LLM methods that require costly runtime inference, our framework precomputes semantic embeddings offline and incorporates them into the transformer-based sequential model \cite{yeh2025treasure}. This design preserves semantic richness and sequential context while meeting the latency and scalability requirements of real-world financial transaction systems.

%% file: section/methodology.tex
\section{Methodology}
\label{sec:method}

In this section, we first present an overview of our framework. Then we delve into each component with detailed descriptions and examples to showcase our approach.

\subsection{Framework Overview}

\begin{figure}[t]
  \includegraphics[width=\columnwidth]{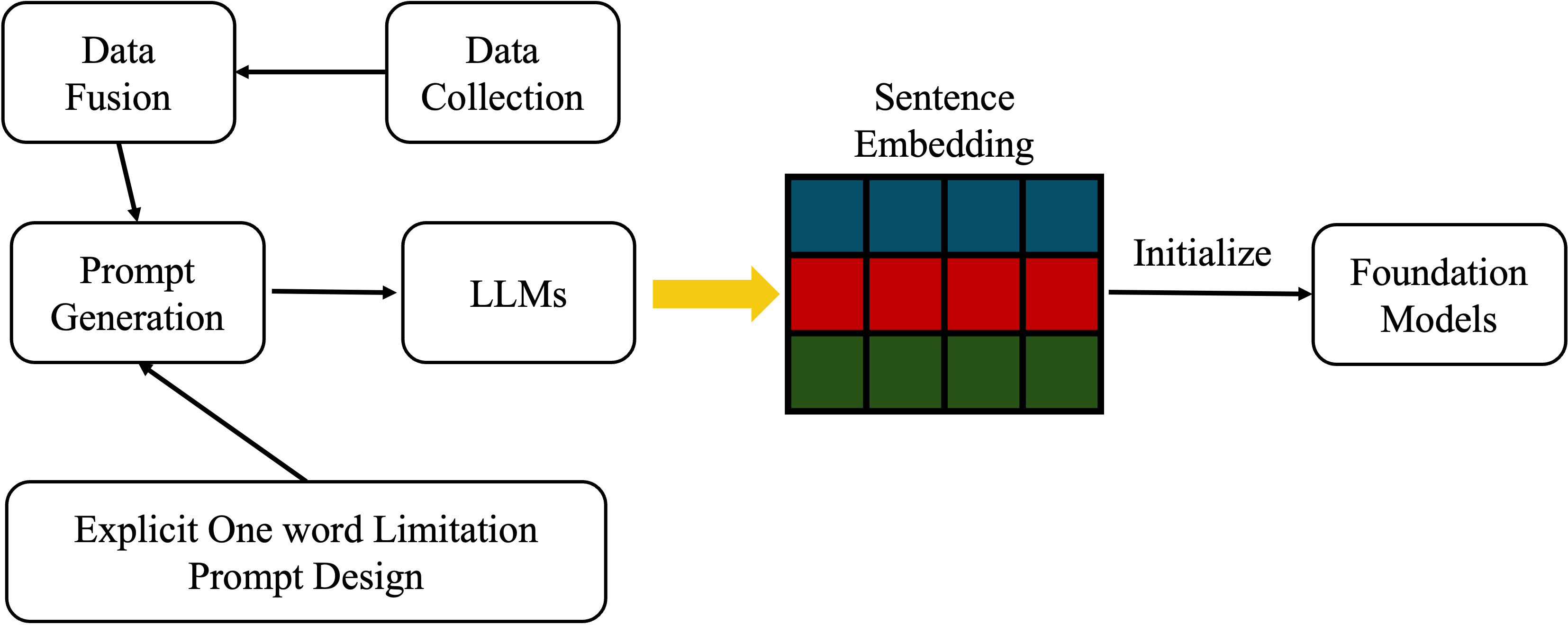}
  \caption{An overview of proposed method.}
  \label{fig:framework}
\end{figure}

Our framework addresses the semantic information loss in traditional foundation models for transaction analysis through a systematic pipeline that leverages LLM capabilities while maintaining production deployment feasibility. Figure~\ref{fig:framework} illustrates the end-to-end architecture, which consists of five interconnected components designed to transform sparse categorical merchant data into semantically rich embeddings.

The pipeline begins with \textbf{Data Collection}, where we gather raw categorical merchant information from transactional datasets, including merchant names, MCC, and location information. This initial step identifies fields amenable to semantic enrichment through external knowledge sources.

\textbf{Data Fusion} enhances the collected categorical data by integrating information from multiple external sources. For MCC, we incorporate official category descriptions, industry classifications, and business type annotations. For merchant names, we augment basic identifiers with business descriptions and its associated MCC. This multi-source approach addresses the sparsity inherent in raw transactional data.

The \textbf{Prompt Generation} component transforms the enriched categorical data into structured natural language inputs optimized for LLM processing. Central to this stage is our \textit{Explicit One-word Limitation Prompt Design} principle, which ensures consistent and focused semantic representations across different LLM architectures. This design choice prevents verbose or inconsistent outputs that could introduce noise in the embedding generation process.

\textbf{LLM Processing} converts the structured prompts into high-dimensional sentence embeddings that capture semantic relationships and contextual knowledge. We extract these embeddings from the final hidden layer representations, specifically targeting the last non-padding token to ensure consistent dimensionality and semantic focus.

Finally, the generated \textbf{Sentence Embeddings} serve as semantic initializations for categorical field embeddings in the foundation models \cite{yeh2025treasure}. This initialization strategy allows lightweight transaction models to inherit rich semantic understanding from LLMs while maintaining computational efficiency through subsequent task-specific fine-tuning.

\subsection{Data Preprocessing and Prompt Generation}
\label{sec:data_preprocessing}

Our data preprocessing pipeline addresses the challenge of transforming sparse categorical merchant information into semantically rich representations suitable for LLM processing. This process consists of two primary stages: multi-source data fusion and structured prompt generation.

\subsubsection{Multi-source Data Fusion}

We develop a comprehensive data integration pipeline that enriches categorical fields through external knowledge sources. Initial experiments with simple, single-source prompts (e.g., "provide the embedding of MCC 5044") yielded suboptimal results, as they failed to capture the rich semantic relationships between related categorical features. This motivated our multi-source fusion approach that leverages contextual information from multiple related fields.
The process begins with structured data extraction from multiple sources, including official MCC documentation \cite{ISO18245-2023}, and internal merchant and geographical information databases. For MCC, we integrate official category descriptions, business type classifications, and related merchant examples. Location data is augmented with economic indicators, demographic information, and region-specific financial patterns.

\subsubsection{Explicit One-word Limitation Prompt Design}

To ensure consistency and reduce noise in LLM outputs, we implement an explicit one-word limitation principle in our prompt design. This approach addresses the challenge of handling verbose or inconsistent LLM responses that could introduce noise in the embedding generation process. Our prompt templates are structured to provide comprehensive context while constraining output format. We design field-specific templates that incorporate relevant contextual information while maintaining semantic coherence across different data types.

\subsection{LLM-based Sentence Embedding Generation}
\label{sec:llm_embedding}

We generate semantically rich embeddings for three primary categorical fields: location information, MCC, and merchant identifiers. Each field type requires specialized prompt engineering to capture domain-specific semantic relationships.

\subsubsection{Location Embedding Generation}

Location prompts incorporate geographical, economic, and regulatory context relevant to financial transactions. We design prompts that capture economic indicators, and regional financial attibutes.

\begin{listing}[htp]
\begin{minted}[breaklines,xleftmargin=2em,linenos,fontsize=\scriptsize]{text}
Input: "UNITED STATES OF AMERICA, New York"
Prompt: "Represent the following location in the context of financial transactions, and economic indicators: New York, USA. 
Consider state-specific economic trends, population demographics, major industries, and financial regulations."
\end{minted}
\caption{Location embedding prompt example for state-level data}
\label{list:location_state}
\end{listing}

\subsubsection{MCC Embedding Generation}

MCC prompts leverage official category descriptions and business characteristics to generate semantically meaningful representations. We provide both basic and enriched versions depending on available contextual information. For a comprehensive MCC representations, we incorporate detailed business descriptions, included categories, and similar merchant types:

\begin{listing}[htp]
\begin{minted}[breaklines,xleftmargin=2em,linenos,fontsize=\scriptsize]{text}
Input: MCC "5044"
Prompt: "The MCC 5044, titled 'Photographic, Photocopy, Microfilm Equipment and Supplies', serves business-to-business distributors of office and photographic equipment including film, cash registers, photocopy machines, and microfilm machines. Similar categories include 5021 (Office Furniture), 5045 (Computer Equipment), and 5943 (Stationery Stores).
Please provide the embedding of MCC 5044."
\end{minted}
\caption{Enriched MCC embedding prompt with contextual information}
\label{list:mcc_enriched}
\end{listing}

\subsubsection{Merchant Embedding Generation}

Merchant prompts combine location, MCC category, and business name information to create comprehensive merchant representations:

\begin{listing}[htp]
\begin{minted}[breaklines,xleftmargin=2em,linenos,fontsize=\scriptsize]{text}
Input: "365 MARKET 888 432-3299"
Prompt: "The merchant '365 MARKET 888 432-3299' is located in Troy, Michigan, USA. It belongs to MCC category 5814 'Fast Food Restaurants', which serves prepared food and beverages for on-premises or carry-out consumption.
Please provide the merchant embedding."
\end{minted}
\caption{Merchant embedding prompt combining location and category context}
\label{list:merchant}
\end{listing}

\subsubsection{Embedding Extraction and Processing}

We evaluate multiple open-source LLMs for generating sentence embeddings. The embedding extraction process involves extracting embeddings from the final hidden layer, specifically targeting the representation of the last non-padding token.

\subsection{Foundation Model Initialization and Training}
\label{sec:model_training}

The core innovation of our framework lies in the semantic initialization strategy, where LLM-generated embeddings serve as starting points for training our foundation model \cite{yeh2025treasure}. By precomputing embeddings for selected attributes (merchant name, MCC, location) offline, we decouple the semantic enrichment phase from the inference pipeline. Once these embeddings initialize the model's embedding layers, the architecture and training procedure remain identical to the original tabular foundation model framework—we refer readers to \cite{yeh2025treasure} for complete architectural specifications and training details.

\subsubsection{Embedding Initialization Strategy}

We initialize categorical field embedding layers with corresponding LLM-generated representations. This initialization provides several advantages: (1) immediate access to semantic relationships encoded by LLMs, (2) reduced training time for achieving semantic coherence, and (3) improved generalization to unseen categorical values with similar semantic properties.

\subsubsection{Multi-task Learning Framework}

Our training framework supports multiple downstream objectives simultaneously, including transaction metric assessment and transaction prediction. By jointly optimizing for these objectives, the framework enables the model to learn versatile representations that generalize across various tasks. This approach balances semantic richness with task-specific performance, enabling foundation models to leverage LLM capabilities while maintaining computational efficiency for production deployment.

%% file: section/experiments.tex
\section{Experiments}
\label{sec:experiments}

\begin{table*}[t]
\centering
\caption{Performance comparison of various LLM architectures (\textbf{Model}) and embedding initialization methods (\textbf{Emb.}) on transaction prediction tasks. “Vanilla” denotes the original foundation model. Green-highlighted cells indicate settings where the model outperforms the vanilla baseline.}
\label{tab:main_res}
\small
\begin{tabular}{lcllllllll}
\toprule
\multirow{2}{*}{\textbf{Model}} & \multirow{2}{*}{\textbf{Emb.}} & \multicolumn{2}{c}{\textbf{Next Amount}} & \multicolumn{2}{c}{\textbf{Next MCC}} & \multicolumn{2}{c}{\textbf{Next City}} & \multicolumn{2}{c}{\textbf{Next Merchant}} \\
 & & \multicolumn{1}{c}{\textbf{MAE}} & \multicolumn{1}{c}{\textbf{sMAPE}} & \multicolumn{1}{c}{\textbf{Acc}} & \multicolumn{1}{c}{\textbf{F1}} & \multicolumn{1}{c}{\textbf{Acc}} & \multicolumn{1}{c}{\textbf{F1}} & \multicolumn{1}{c}{\textbf{Acc}} & \multicolumn{1}{c}{\textbf{F1}} \\
\midrule
\multicolumn{2}{c}{Vanilla} & 37.0430 & 0.3952 & 0.4107 & 0.1118 & 0.8454 & 0.6721 & 0.0760 & 0.0037 \\
\midrule
Llama2-7b & \multirow{4}{*}{MCC} 
 & \cellcolor{green!20}36.7474
 & 0.3954
 & \cellcolor{green!20}0.4134
 & \cellcolor{green!20}0.1143
 & 0.8443
 & 0.6683
 & \cellcolor{green!20}0.0913
 & \cellcolor{green!20}0.0099 \\
Llama2-13b & &
 37.2129
 & 0.3978
 & \cellcolor{green!20}0.4118
 & \cellcolor{green!20}0.1148
 & 0.8436
 & 0.6652
 & \cellcolor{green!20}0.0905
 & \cellcolor{green!20}0.0103 \\
Llama3-8b & &
 37.5144
 & 0.4010
 & \cellcolor{green!20}0.4132
 & \cellcolor{green!20}0.1131
 & \cellcolor{green!20}0.8456
 & \cellcolor{green!20}0.6803
 & \cellcolor{green!20}0.0960
 & \cellcolor{green!20}0.0103 \\
Mistral-7b & &
 \cellcolor{green!20}37.0173
 & 0.3961
 & \cellcolor{green!20}0.4152
 & 0.1108
 & 0.8449
 & 0.6623
 & \cellcolor{green!20}0.0934
 & \cellcolor{green!20}0.0100 \\
\midrule
Llama2-7b & \multirow{4}{*}{Merchant}
 & \cellcolor{green!20}36.9498
 & \cellcolor{green!20}0.3949
 & \cellcolor{green!20}0.4119
 & 0.1054
 & 0.8442
 & 0.6625
 & \cellcolor{green!20}0.0938
 & \cellcolor{green!20}0.0098 \\
Llama2-13b & &
 \cellcolor{green!20}37.0119
 & 0.3962
 & \cellcolor{green!20}0.4123
 & 0.1014
 & 0.8437
 & 0.6597
 & \cellcolor{green!20}0.0898
 & \cellcolor{green!20}0.0085 \\
Llama3-8b & &
 37.0511
 & 0.3968
 & \cellcolor{green!20}0.4115
 & 0.1102
 & 0.8434
 & 0.6606
 & \cellcolor{green!20}0.0991
 & \cellcolor{green!20}0.0098 \\
Mistral-7b & &
 \cellcolor{green!20}36.8613
 & 0.3957
 & \cellcolor{green!20}0.4148
 & \cellcolor{green!20}0.1141
 & \cellcolor{green!20}0.8456
 & 0.6649
 & \cellcolor{green!20}0.0930
 & \cellcolor{green!20}0.0095 \\
\midrule
Llama2-7b & \multirow{4}{*}{\begin{tabular}[c]{@{}c@{}}MCC\\ Merchant\end{tabular}}
 & \cellcolor{green!20}37.0336
 & 0.3986
 & \cellcolor{green!20}0.4145
 & \cellcolor{green!20}0.1177
 & 0.8452
 & 0.6603
 & \cellcolor{green!20}0.0929
 & \cellcolor{green!20}0.0103 \\
Llama2-13b & &
 \cellcolor{green!20}36.8841
 & \cellcolor{green!20}0.3932
 & \cellcolor{green!20}0.4154
 & \cellcolor{green!20}0.1162
 & 0.8451
 & 0.6536
 & \cellcolor{green!20}0.0946
 & \cellcolor{green!20}0.0100 \\
Llama3-8b & &
 \cellcolor{green!20}36.8522
 & 0.3957
 & \cellcolor{green!20}0.4168
 & \cellcolor{green!20}0.1202
 & \cellcolor{green!20}0.8458
 & 0.6636
 & \cellcolor{green!20}0.0994
 & \cellcolor{green!20}0.0110 \\
Mistral-7b & &
 37.1708
 & 0.3983
 & \cellcolor{green!20}0.4126
 & 0.1052
 & 0.8433
 & 0.6462
 & \cellcolor{green!20}0.0901
 & \cellcolor{green!20}0.0083 \\
\midrule
Llama2-7b & \multirow{4}{*}{\begin{tabular}[c]{@{}c@{}}State\\ City\end{tabular}}
 & 37.3200
 & 0.4025
 & \cellcolor{green!20}0.4163
 & \cellcolor{green!20}0.1157
 & 0.8450
 & 0.6642
 & \cellcolor{green!20}0.0959
 & \cellcolor{green!20}0.0101 \\
Llama2-13b & &
 \cellcolor{green!20}36.7465
 & \cellcolor{green!20}0.3935
 & \cellcolor{green!20}0.4151
 & \cellcolor{green!20}0.1124
 & \cellcolor{green!20}0.8459
 & 0.6646
 & \cellcolor{green!20}0.0951
 & \cellcolor{green!20}0.0120 \\
Llama3-8b & &
 \cellcolor{green!20}36.7652
 & 0.3952
 & \cellcolor{green!20}0.4146
 & \cellcolor{green!20}0.1187
 & \cellcolor{green!20}0.8451
 & 0.6607
 & \cellcolor{green!20}0.0947
 & \cellcolor{green!20}0.0106 \\
Mistral-7b & &
 \cellcolor{green!20}36.9257
 & 0.3972
 & \cellcolor{green!20}0.4125
 & 0.1076
 & 0.8437
 & 0.6577
 & \cellcolor{green!20}0.0904
 & \cellcolor{green!20}0.0090 \\
\midrule
Llama2-7b & \multirow{4}{*}{All Fields}
 & \cellcolor{green!20}37.0218
 & 0.3977
 & \cellcolor{green!20}0.4140
 & \cellcolor{green!20}0.1178
 & \cellcolor{green!20}0.8462
 & 0.6683
 & \cellcolor{green!20}0.0942
 & \cellcolor{green!20}0.0098 \\
Llama2-13b & &
 37.1253
 & 0.4003
 & \cellcolor{green!20}0.4152
 & \cellcolor{green!20}0.1184
 & \cellcolor{green!20}0.8454
 & 0.6651
 & \cellcolor{green!20}0.0991
 & \cellcolor{green!20}0.0115 \\
Llama3-8b & &
 \cellcolor{green!20}36.8128
 & \cellcolor{green!20}0.3927
 & \cellcolor{green!20}0.4155
 & \cellcolor{green!20}0.1208
 & \cellcolor{green!20}0.8455
 & 0.6716
 & \cellcolor{green!20}0.0979
 & \cellcolor{green!20}0.0110 \\
Mistral-7b & &
 \cellcolor{green!20}36.8761
 & 0.3955
 & \cellcolor{green!20}0.4108
 & 0.0960
 & 0.8434
 & 0.6398
 & \cellcolor{green!20}0.0979
 & \cellcolor{green!20}0.0074 \\
\bottomrule
\end{tabular}
\end{table*}

In this section, we present the results of comprehensive experiments to evaluate the effectiveness of our LLM-based semantic initialization framework across multiple transaction understanding tasks. Our evaluation encompasses both the quality of generated embeddings and their impact on downstream financial applications. We use four different LLMs for embedding initialization:  LLama2-7b and LLama2-13b \cite{touvron2023llama}, Llama3-8b \cite{grattafiori2024llama}, and Mistral-7b \cite{jiang2024mistral}.

\subsection{Dataset}
\label{sec:dataset}

We conduct experiments using one billion transaction records from January 2022 to December 2023, covering diverse merchant categories, user behaviors, and transaction patterns. Transactions from the first 20 months are used for model training, the 21st month serves as the validation data, and transaction from the final 3 months is used as the test data. The structured dataset, enriched with contextual information from our preprocessing pipeline, offers a robust testbed for evaluating semantic embeddings across varied merchants, locations, and transaction contexts.

\subsection{Evaluation Tasks}
\label{sec:eval_tasks}

Our evaluation focuses on five tasks that assess different aspects of semantic understanding:

\textbf{Next Amount Prediction}: Regression task forecasting a user’s next transaction amount, capturing spending patterns and merchant pricing.

\textbf{Next MCC Prediction}: Multi-class classification of the next transaction’s merchant category code, evaluating business-type relationships.

\textbf{Next Location Prediction}: Classification of the next transaction’s location (City), assessing spatial patterns and preferences.

\textbf{Next Merchant Prediction}: Multi-class classification of the next merchant, requiring detailed understanding of merchants and user preferences.

\textbf{Transaction Metrics Assessment}: Binary classification task for identifying anomalous transactions based on merchant characteristics, transaction patterns, and contextual features. This task requires real-time inference capabilities and high precision to minimize false classifications while maintaining operational efficiency.

These tasks collectively evaluate the framework's ability to capture semantic relationships across different categorical fields and their impact on transaction understanding.

\subsection{Evaluation Metrics}
\label{sec:metrics}

We employ task-specific metrics to comprehensively assess performance. The reported metrics include mean absolute error (MAE) and symmetric mean absolute percentage error (sMAPE) for amount prediction, as well as accuracy (Acc) and F1-score for classification.

\textbf{Transaction Metrics Assessment}: An in-house performance metric designed for evaluating transaction understanding capabilities. Due to the sensitive nature of this task, detailed performance figures cannot be disclosed. To present these results, we compute the performance ratio between the evaluated model and the currently deployed system, termed Relative Improvement (RI):

\begin{equation} \text{RI} = \frac{S_{\text{evaluated}} - S_{\text{baseline}}}{S_{\text{baseline}}} \times 100\% \end{equation}

where $S_{\text{evaluated}}$ represents the performance score of the evaluated model and $S_{\text{baseline}}$ represents the performance score of the currently deployed baseline system. A positive RI indicates performance improvement over the baseline, while a negative RI indicates performance degradation. For instance, RI=+50\% indicates that the evaluated model achieves 50\% superior performance compared to the currently deployed system in transaction pattern recognition. This normalized approach enables meaningful performance comparison while adhering to confidentiality requirements for production financial systems.

\subsection{Baseline Comparisons}
\label{sec:baselines}

We compare our approach against a vanilla foundation model \cite{yeh2025treasure} baseline that uses traditional ID-based categorical representations without LLM-based semantic embeddings. This baseline represents the current state-of-practice in transaction modeling and provides a direct assessment of semantic initialization benefits.

\begin{table}[t]
\centering
\caption{Transaction Metrics Assessment: Relative Improvement (RI) across different embedding initialization strategies}
\small
\label{tab:transaction_metrics}
\begin{tabular}{l|c|c|c}
\toprule
\textbf{Model} & \textbf{MCC + Merchant} & \textbf{Location} & \textbf{All Fields} \\
\midrule
Vanilla & $S_{\text{baseline}}$ & $S_{\text{baseline}}$ & $S_{\text{baseline}}$ \\
Llama2-7b & -0.40\% & +2.85\% & +3.72\% \\
Llama2-13b & +2.66\% & +1.77\% & +2.92\% \\
Llama3-8b & +0.37\% & +2.78\% & +3.32\% \\
Mistral-7b & +0.83\% & +2.89\% & +3.93\% \\
\bottomrule
\end{tabular}
\end{table}

\subsection{Results and Analysis}
\label{sec:results}

Table~\ref{tab:main_res} presents comprehensive results comparing different LLM architectures and embedding initialization strategies across transaction prediction tasks. The numbers reported are averages over three runs. Grey-highlighted cells indicate configurations that outperform the vanilla baseline, revealing systematic patterns in semantic initialization effectiveness. 

\textbf{Embedding Initialization Strategy Analysis} Our experimental results reveal distinct performance characteristics across initialization strategies. Single-field initialization strategies exhibit more specialized benefits. While MCC-only and merchant-only approaches both improve performance on MCC and merchant prediction tasks, combining them further boosts results. Location-focused initialization presents mixed but generally positive outcomes. It demonstrates a boost in the next location prediction task, where other initialization strategies fail to show superiority. Comprehensive all-fields initialization yields strong performance across most configurations, though some architectures exhibit reduced performance on specific metrics, indicating potential optimization challenges when integrating multiple embedding types simultaneously.

\textbf{Task-Specific Performance Analysis} Amount prediction tasks benefit from all initialization strategies except the MCC-only approach, suggesting that geographic and holistic semantic understanding significantly enhance spending pattern prediction, while MCC information alone offers limited value. MCC/merchant prediction demonstrates the most consistent improvements across all initialization strategies, with relative performance gains observed in 82\%/100\% of experimental configurations. This high success rate indicates that LLM-generated embeddings effectively capture semantic relationships between merchant categories and merchants. The universal nature of these gains across all configurations underscores the critical importance of semantic merchant understanding for fine-grained prediction tasks. Location prediction tasks show more modest improvements, with the strongest gains achieved through all-fields initialization. This can be attributed to the relative simplicity of the task, as location information in a transaction sequence tends to be consistent, and the vanilla model already demonstrates good performance.

\textbf{Architecture-Specific Performance Characteristics} Cross-architectural analysis reveals distinct performance profiles. Llama3-8b emerges as the most versatile architecture, consistently achieving top-tier performance across multiple initialization strategies and tasks. The model's ability to effectively leverage semantic relationships across different field combinations suggests superior architectural suitability for transaction understanding applications.

\textbf{Transaction Metrics Assessment} Table~\ref{tab:transaction_metrics} presents Relative Improvement (RI) results for our proprietary transaction metrics assessment. Nearly all initialization strategies demonstrate superior performance, showcasing the effectiveness of the proposed method in real-world applications.

%% file: section/conclusion.tex
\section{Conclusion}
\label{sec:conclusion}

In this work, we addressed the limitations of traditional transaction analysis models that rely on index-based representations for categorical merchant fields, leading to a loss of valuable semantic information. Building upon the tabular foundation model framework in \cite{yeh2025treasure}, we propose a hybrid approach that integrates LLM-generated embeddings as semantic initializations. This integration preserves the computational efficiency and scalability of the original architecture while enriching it with semantic understanding from large language models. By precomputing LLM embeddings offline and using them to initialize embedding layers, we eliminate runtime LLM inference costs while retaining semantic richness. Our framework further leverages multi-source data fusion and a one-word constraint principle to ensure consistency and robustness in embedding generation. Through systematic noise filtering and context-aware data enrichment, we enhance data quality and model reliability. Empirical results on large-scale transaction datasets validate the effectiveness of our framework, demonstrating significant improvements across multiple transaction understanding tasks while maintaining the production-ready efficiency of the baseline model \cite{yeh2025treasure}.

%% file: section/limitations.tex
\section*{Limitations}
While our framework demonstrates significant improvements across multiple transaction understanding tasks, several limitations warrant acknowledgment and suggest directions for future research.

\textbf{Prompt Engineering Sophistication}: Our Explicit One-word Limitation Prompt Design, while effective for consistency, represents a relatively simple approach to prompt engineering. More sophisticated prompt optimization techniques and domain-specific fine-tuning of LLMs for financial contexts could potentially enhance semantic representation quality.

\textbf{Model Coverage}: Our evaluation focuses on established LLM architectures and does not include the most recent state-of-the-art models or specialized sentence embedding models such as NV-Embed~\cite{lee2024nv} and Qwen3-embedding~\cite{zhang2025qwen3}. Testing with these advanced models could potentially yield further performance improvements and provide additional insights into architectural suitability for financial domain applications.

\textbf{Categorical Field Scope}: Our framework currently addresses MCC, merchant, and location embeddings, but does not extend to other potentially valuable categorical fields such as transaction channels, payment methods, or temporal patterns. The generalizability of our approach to these additional fields remains to be validated.

\textbf{Static Embedding Approach}: The current framework generates embeddings offline and does not account for evolving merchant characteristics, seasonal business patterns, or dynamic market conditions. This static approach may limit the framework's ability to capture temporal semantic changes in transaction contexts.

These limitations highlight opportunities for future research while acknowledging the scope and constraints of our current contribution to LLM-based semantic enhancement in financial transaction understanding.

%% file: main.bbl
\begin{thebibliography}{24}
\providecommand{\natexlab}[1]{#1}

\bibitem[{ISO(2023)}]{ISO18245-2023}
 2023.
\newblock \href {https://www.iso.org/standard/79450.html} {Retail financial services — merchant category codes}.
\newblock International Standard ISO 18245:2023, International Organization for Standardization.

\bibitem[{Alayrac et~al.(2022)Alayrac, Donahue, Luc, Miech, Barr, Hasson, Lenc, Mensch, Millican, Reynolds et~al.}]{alayrac2022flamingo}
Jean-Baptiste Alayrac, Jeff Donahue, Pauline Luc, Antoine Miech, Iain Barr, Yana Hasson, Karel Lenc, Arthur Mensch, Katherine Millican, Malcolm Reynolds, and 1 others. 2022.
\newblock Flamingo: a visual language model for few-shot learning.
\newblock \emph{Advances in neural information processing systems}, 35:23716--23736.

\bibitem[{Awais et~al.(2025)Awais, Naseer, Khan, Anwer, Cholakkal, Shah, Yang, and Khan}]{awais2025foundation}
Muhammad Awais, Muzammal Naseer, Salman Khan, Rao~Muhammad Anwer, Hisham Cholakkal, Mubarak Shah, Ming-Hsuan Yang, and Fahad~Shahbaz Khan. 2025.
\newblock Foundation models defining a new era in vision: a survey and outlook.
\newblock \emph{IEEE Transactions on Pattern Analysis and Machine Intelligence}.

\bibitem[{Brown et~al.(2020)Brown, Mann, Ryder, Subbiah, Kaplan, Dhariwal, Neelakantan, Shyam, Sasaki, Askell et~al.}]{brown2020language}
Tom Brown, Benjamin Mann, Nick Ryder, Melanie Subbiah, Jared~D Kaplan, Prafulla Dhariwal, Arvind Neelakantan, Pranav Shyam, Girish Sasaki, Amanda Askell, and 1 others. 2020.
\newblock Language models are few-shot learners.
\newblock \emph{Advances in neural information processing systems}, 33:1877--1901.

\bibitem[{Devlin et~al.(2018)Devlin, Chang, Lee, and Toutanova}]{devlin2018bert}
Jacob Devlin, Ming-Wei Chang, Kenton Lee, and Kristina Toutanova. 2018.
\newblock Bert: Pre-training of deep bidirectional transformers for language understanding.
\newblock \emph{arXiv preprint arXiv:1810.04805}.

\bibitem[{Dosovitskiy et~al.(2020)Dosovitskiy, Beyer, Kolesnikov, Weissenborn, Zhai, Unterthiner, Dehghani, Minderer, Heigold, Gelly et~al.}]{dosovitskiy2020image}
Alexey Dosovitskiy, Lucas Beyer, Alexander Kolesnikov, Dirk Weissenborn, Xiaohua Zhai, Thomas Unterthiner, Mostafa Dehghani, Matthias Minderer, Georg Heigold, Sylvain Gelly, and 1 others. 2020.
\newblock An image is worth 16x16 words: Transformers for image recognition at scale.
\newblock \emph{arXiv preprint arXiv:2010.11929}.

\bibitem[{Grattafiori et~al.(2024)Grattafiori, Dubey, Jauhri, Pandey, Kadian, Al-Dahle, Letman, Mathur, Schelten, Vaughan et~al.}]{grattafiori2024llama}
Aaron Grattafiori, Abhimanyu Dubey, Abhinav Jauhri, Abhinav Pandey, Abhishek Kadian, Ahmad Al-Dahle, Aiesha Letman, Akhil Mathur, Alan Schelten, Alex Vaughan, and 1 others. 2024.
\newblock The llama 3 herd of models.
\newblock \emph{arXiv preprint arXiv:2407.21783}.

\bibitem[{He et~al.(2025)He, Liu, Zhang, Ma, and Chua}]{he2025large}
Yingzhi He, Xiaohao Liu, An~Zhang, Yunshan Ma, and Tat-Seng Chua. 2025.
\newblock Llm2rec: Large language models are powerful embedding models for sequential recommendation.
\newblock \emph{arXiv preprint arXiv:2506.21579}.

\bibitem[{Hollmann et~al.(2022)Hollmann, M{\"u}ller, Eggensperger, and Hutter}]{hollmann2022tabpfn}
Noah Hollmann, Samuel M{\"u}ller, Katharina Eggensperger, and Frank Hutter. 2022.
\newblock Tabpfn: A transformer that solves small tabular classification problems in a second.
\newblock \emph{arXiv preprint arXiv:2207.01848}.

\bibitem[{Huang et~al.(2024)Huang, Yu, Xie, Zhang, Yao, and McAuley}]{huang2024foundation}
Chengkai Huang, Tong Yu, Kaige Xie, Shuai Zhang, Lina Yao, and Julian McAuley. 2024.
\newblock Foundation models for recommender systems: A survey and new perspectives.
\newblock \emph{arXiv preprint arXiv:2402.11143}.

\bibitem[{Jiang et~al.(2024{\natexlab{a}})Jiang, Sablayrolles, Mensch, Bamford, Chaplot, Casas, Bressand, Lengyel, Lample, Saulnier et~al.}]{jiang2024mistral}
AQ~Jiang, A~Sablayrolles, A~Mensch, C~Bamford, DS~Chaplot, Ddl Casas, F~Bressand, G~Lengyel, G~Lample, L~Saulnier, and 1 others. 2024{\natexlab{a}}.
\newblock Mistral 7b. arxiv 2023.
\newblock \emph{arXiv preprint arXiv:2310.06825}.

\bibitem[{Jiang et~al.(2024{\natexlab{b}})Jiang, Huang, Luan, Wang, and Zhuang}]{jiang2024scaling}
Ting Jiang, Shaohan Huang, Zhongzhi Luan, Deqing Wang, and Fuzhen Zhuang. 2024{\natexlab{b}}.
\newblock Scaling sentence embeddings with large language models.
\newblock In \emph{Findings of the Association for Computational Linguistics: EMNLP 2024}, pages 3182--3196.

\bibitem[{Lee et~al.(2024)Lee, Roy, Xu, Raiman, Shoeybi, Catanzaro, and Ping}]{lee2024nv}
Chankyu Lee, Rajarshi Roy, Mengyao Xu, Jonathan Raiman, Mohammad Shoeybi, Bryan Catanzaro, and Wei Ping. 2024.
\newblock Nv-embed: Improved techniques for training llms as generalist embedding models.
\newblock \emph{arXiv preprint arXiv:2405.17428}.

\bibitem[{Ramesh et~al.(2021)Ramesh, Pavlov, Goh, Gray, Voss, Radford, Chen, and Sutskever}]{ramesh2021zero}
Aditya Ramesh, Mikhail Pavlov, Gabriel Goh, Scott Gray, Chelsea Voss, Alec Radford, Mark Chen, and Ilya Sutskever. 2021.
\newblock Zero-shot text-to-image generation.
\newblock In \emph{International conference on machine learning}, pages 8821--8831. Pmlr.

\bibitem[{Schafer et~al.(2007)Schafer, Frankowski, Herlocker, and Sen}]{schafer2007collaborative}
J~Ben Schafer, Dan Frankowski, Jon Herlocker, and Shilad Sen. 2007.
\newblock Collaborative filtering recommender systems.
\newblock In \emph{The adaptive web: methods and strategies of web personalization}, pages 291--324. Springer.

\bibitem[{Skalski et~al.(2023)Skalski, Sutton, Burrell, Perez, and Wong}]{skalski2023towards}
Piotr Skalski, David Sutton, Stuart Burrell, Iker Perez, and Jason Wong. 2023.
\newblock Towards a foundation purchasing model: Pretrained generative autoregression on transaction sequences.
\newblock In \emph{Proceedings of the Fourth ACM International Conference on AI in Finance}, pages 141--149.

\bibitem[{Su et~al.(2024)Su, Wang, Ye, Zhou, Zhang, Chen, Zhu, Wang, Xu, Chen et~al.}]{su2024tablegpt2}
Aofeng Su, Aowen Wang, Chao Ye, Chen Zhou, Ga~Zhang, Gang Chen, Guangcheng Zhu, Haobo Wang, Haokai Xu, Hao Chen, and 1 others. 2024.
\newblock Tablegpt2: A large multimodal model with tabular data integration.
\newblock \emph{arXiv preprint arXiv:2411.02059}.

\bibitem[{Touvron et~al.(2023)Touvron, Martin, Stone, Albert, Almahairi, Babaei, Bashlykov, Batra, Bhargava, Bhosale et~al.}]{touvron2023llama}
Hugo Touvron, Louis Martin, Kevin Stone, Peter Albert, Amjad Almahairi, Yasmine Babaei, Nikolay Bashlykov, Soumya Batra, Prajjwal Bhargava, Shruti Bhosale, and 1 others. 2023.
\newblock Llama 2: Open foundation and fine-tuned chat models.
\newblock \emph{arXiv preprint arXiv:2307.09288}.

\bibitem[{Van~Breugel and Van Der~Schaar(2024)}]{van2024tabular}
Boris Van~Breugel and Mihaela Van Der~Schaar. 2024.
\newblock Why tabular foundation models should be a research priority.
\newblock \emph{arXiv preprint arXiv:2405.01147}.

\bibitem[{Yang et~al.(2023)Yang, Wang, Liu, Wu, and Liu}]{yang2023unitabe}
Yazheng Yang, Yuqi Wang, Guang Liu, Ledell Wu, and Qi~Liu. 2023.
\newblock Unitabe: A universal pretraining protocol for tabular foundation model in data science.
\newblock \emph{arXiv preprint arXiv:2307.09249}.

\bibitem[{Yeh et~al.(2025)Yeh, Saini, Dai, Fan, Jain, Fan, Sun, Wang, Pan, Dou, Rakesh, Wang, Zheng, and Das}]{yeh2025treasure}
Chin-Chia~Michael Yeh, Uday~Singh Saini, Xin Dai, Xiran Fan, Shubham Jain, Yujie Fan, Jiarui Sun, Junpeng Wang, Menghai Pan, Yuzhong Dou, Yingtong~Chen, Vineeth Rakesh, Liang Wang, Yan Zheng, and Mahashweta Das. 2025.
\newblock {TREASURE}: A transformer-based foundation model for high-volume transaction understanding.
\newblock \emph{arXiv preprint arXiv:2511.19693}.

\bibitem[{Zhang et~al.(2023{\natexlab{a}})Zhang, Wang, Dai, Jain, Wang, Fan, Yeh, Zheng, Zhuang, and Zhang}]{zhang2023fata}
Dongyu Zhang, Liang Wang, Xin Dai, Shubham Jain, Junpeng Wang, Yujie Fan, Chin-Chia~Michael Yeh, Yan Zheng, Zhongfang Zhuang, and Wei Zhang. 2023{\natexlab{a}}.
\newblock Fata-trans: Field and time-aware transformer for sequential tabular data.
\newblock In \emph{Proceedings of the 32nd ACM International Conference on Information and Knowledge Management}, pages 3247--3256.

\bibitem[{Zhang et~al.(2023{\natexlab{b}})Zhang, Wen, Zheng, Xu, and Bian}]{zhang2023towards}
Han Zhang, Xumeng Wen, Shun Zheng, Wei Xu, and Jiang Bian. 2023{\natexlab{b}}.
\newblock Towards foundation models for learning on tabular data.

\bibitem[{Zhang et~al.(2025)Zhang, Li, Long, Zhang, Lin, Yang, Xie, Yang, Liu, Lin et~al.}]{zhang2025qwen3}
Yanzhao Zhang, Mingxin Li, Dingkun Long, Xin Zhang, Huan Lin, Baosong Yang, Pengjun Xie, An~Yang, Dayiheng Liu, Junyang Lin, and 1 others. 2025.
\newblock Qwen3 embedding: Advancing text embedding and reranking through foundation models.
\newblock \emph{arXiv preprint arXiv:2506.05176}.

\end{thebibliography}
